# Multiple-Vehicle Tracking in the Highway Using Appearance Model and Visual Object Tracking

Fateme Bafghi
dept. Artificial Intelligence (of University of Isfahan)
Isfahan, Iran
fatemebfg@gmail.com

Bijan Shoushtarian
dept. Artificial Intelligence (of University of Isfahan)
Isfahan, Iran
shoushtarian@eng.ui.ac.ir

*Abstract*—In recent decades, due to the groundbreaking improvements in machine vision, many daily tasks are performed by computers. One of these tasks is multiple-vehicle tracking, which is widely used in different areas such as video surveillance and traffic monitoring. This paper focuses on introducing an efficient novel approach with acceptable accuracy. This is achieved through an efficient appearance and motion model based on the features extracted from each object. For this purpose, two different approaches have been used to extract features, i.e. features extracted from a deep neural network, and traditional features. Then the results from these two approaches are compared with state-of-the-art trackers. The results are obtained by executing the methods on the UA-DETRACK benchmark. The first method led to 58.9% accuracy while the second method caused up to 15.9%. The proposed methods can still be improved by extracting more distinguishable features.

## I. INTRODUCTION

Multiple-object tracking is one of the most practical and challenging fields in machine vision and image processing. Generally, tracking means using spatio-temporal and morphological similarities in adjacent frames to distinguish an object's position as well as assigning a unique ID to it [1]. Tracking is used widely in different fields such as automatic monitoring systems, traffic control, moving systems of autonomous vehicles, etc. The tracking itself is divided into two categories according to the number of tracking objects. These are Single-Object Tracking[1] and Multiple-Object Tracking[2]. In Single-Object Tracking, the system tracks only one object which its first position is initialized at the first time is being tracked but in multiple-object tracking, more than one object is being tracked. These items can be all in one type or different types [2].

For tracking multiple objects different methods have been proposed. These methods can be categorized into online and offline methods. In one of these methods, first, all the objects are extracted by using object detection algorithms and then by defining an appearance and a motion model the prior position of the object is being estimated. The output of the object recognition stage is called the detection response. Joining detection responses makes a tracklet (a small portion of a tracking path) and by joining tracklets a complete track is made. There are various challenges in multiple-object tracking like tracking objects in crowded scenes, tracking tiny objects, tracking opaque objects, as well as occlusion, misdetection, and light changing of the scene.

In this work, an object detection network with high accuracy is used to recognize objects in each frame. To do so, a Mask R-CNN network has been used which guarantees the accuracy of the detection results. In the next step, a graph among the objects of each frame is created. Since the graph may be very complex with numerous nodes and edges, it is pruned by considering an IOU threshold among the objects. Then, each object is tracked by finding its corresponding node in the graph using an appearance and motion model. For defining the appearance model, two different approaches have been utilized. The first approach defines the appearance model based on a linear function of the SIFT features and color histogram for each object. However, the second approach defines the appearance model by using the features of the Mask R-CNN network. Finally, the performances of both approaches are compared and the superior approach is chosen as the result.

The organization of the paper is as follows. Section II is about the literature survey. The proposed technique is detailed in Section III. Section IV explains the proposed method of implementation. Finally, the experimental results and conclusion are detailed in Section V and VI.

## II. A REVIEW ON MULTIPLE-OBJECT TRACKING

Different methods have been proposed to track multiple objects. In these methods, different approaches like machine learning, image processing, feature extraction, and data association have been used. One of these approaches is using graph-based methods. In this method, the tracking problem is modeled as finding the best route in a graph in which the nodes represent the objects in each frame and the edge between two nodes is calculated through a cost function between them.

The graph made from objects is bipartite. A bipartite graph consists of different parts. In each part, there should be no edge between nodes and there should be no edge between a node and itself as well. The other fact that matters is that between two split trajectories there shouldn't be any mutual nodes. In [3] a network flow has been used. The aim of the method in [3] is to move from a beginning node to an end node such that the sum of edge weights is least. In this method, the scene is divided into k areas and then by defining constraints for each area, the problem is formulated. As one of the constraints defined for each area is the number of people that are in that area. Due to a large number of frames and variables, the problem is considered as an NP-hard problem, so by simplifying the constraints, the problem is changed to a linear one. The weakness of the NP-hard problem is the high probability of converging to minimum optima. Finally, by using the k-shortest path algorithm the best trajectory of each node is found.

---
[1] SOT
[2] MOT

In [4] a graph-based approach has been used too, except that it uses dynamic programming instead of linear programming. These methods are more probable to achieve the global optimum.

In [5] three graphs collaborate to solve the tracking problem. These graphs are an appearance graph, a spatio-temporal graph, and an exclusive graph. These graphs are solved through convergence differences algorithms.

Another approach used in multiple-object tracking is called GMCP [6]. In this method, unlike network flow, there is no simplification but the graph is designed like the real world. In fact, all the connections between different nodes, direct or indirect, have been taken into account. This job has some drawbacks because due to the greedy approach it uses, the local optima may be achieved and in addition, the tracking task is executed for each object separately. The biggest drawback of this method is being too slow for multiple consecutive frames. In another work that is popular as IOU tracking [7], only the bounding boxes of an object have been used to estimate an object's future position. By relying on the new high accurate detectors, this method proposes a high-speed tracker with a low computational cost. The drawback of this method is that it just uses the object's position gained from a detector and if in some frames, due to occlusion or the tracker fault, the object is missed, the tracker will not work reliably. So the authors published an update to their method which tried to use the object's visual information alongside the IOU thresholding [8]. In fact, in their new method, they substitute the visual tracking with IOU tracking in a situation when none of the detections fit in the IOU threshold.

Another metric that highly affects tracking results is the features used to make the appearance model. There is a wide range of features highly used in vision fields, features such as HOG[3], color histogram, depth or more complicated features like SIFT, ORB, etc. Each of these features has some advantages and disadvantages and should be used based on the problem. In multiple object tracking, transformation, illumination, occlusion, and even scale will have great effects on the final results. So a more robust feature leads to better results on the tracking. In [9] the HOG features with the bag of video words are used to create the appearance model. The SIFT feature is also a robust feature showing a great performance under the change of transformation, illumination and even occlusion. In [10] only the SIFT features are used to make the appearance model.

Another emerging field which has started to make groundbreaking changes in different machine vision fields is deep learning. A lot of methods have been proposed which have tried to use fully connected deep neural networks such as Rolo which is a combination of a recurrent neural network and Yolo [11]. For detecting objects, the authors have used the Yolo network and then by using the features of each object which have been obtained through a CNN network, the spatial information of each object has been achieved. Then these features with objects and previous frames' information are sent to an LSTM cell to predict the object's position.

In 2018, Zhai [12] used features extracted from a deep network in a single-object tracking. In this work, CNN is used to learn the features of a predefined object and then chooses the best answer among some proposed bounding boxes. In [13] similar approach has been used but instead of a simple CNN network, a Siamese network that was trained on a set of vehicle images has been used.

Another method that has been called DeepSORT[4] is another attempt to merge deep networks with classic multiple-object tracking approaches. This method is an improvement to the SORT method[5]. In the SORT method, the authors used an object detection neural network which is faster R-CNN to extract the objects of each frame. In the prediction phase, they have used a Kalman filter to estimate an object [14, 15]. Since the Kalman filter fails in situations like occlusion, changing viewpoint and velocity changing, the authors tried to strengthen their method by involving the object's features from a deep neural network. This network can be a simple YOLO as in [16] or a re-identification CNN as in [17].

III. THE PROPOSED METHOD

The proposed method in this paper is graph-based too. The general steps are as follows, which will be discussed in details in the next sections:

*1)* Extract objects by using deep neural networks

*2)* Generate a bipartite graph between two consecutive frames

*3)* Extract the features of each object and define the appearance model

*4)* Calculate the weight of the edges between nodes using appearance and motion models

*5)* Find the corresponding node with the current node using graph solving algorithms

*6)* Solving the occlusion problem by defining hypothetical nodes

A. *Extracting Objects in Each Frame*

Extracting the objects of each frame is one of the most important steps of the tracking approach. Mis-defining or missing some objects in a frame can cause the malfunctioning of the tracking system. So it's very important to choose a method that has high accuracy in finding the objects in each frame. Till now a lot of approaches have been proposed to extract the object in an image. These methods are divided into two categories: classic approaches and deep neural network object extractors. Between these categories, neural networks are more accurate in object detection tasks. One of the state-of-the-art networks that is widely used in object detection is Mask R-CNN. Mask R-CNN is designed by Facebook AI researchers. It is an upgrade to faster R-CNN and more accurate than it. In order to have a better vision of Mask R-CNN, it's better to first explain faster R-CNN network architecture and then expand the Mask R-CNN's.

---

[3] Histogram of Oriented Gradient
[4] Simple Online and Real time Tracking with a Deep association metric
[5] Simple Online and Real time Tracking

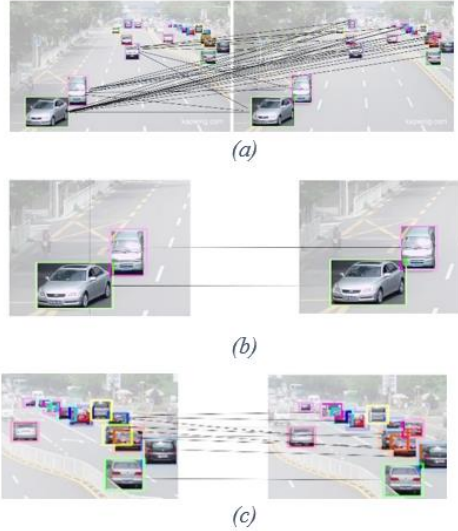

Fig.1. Figure 1-a is the graph between two frames without pruning and Figures 1-b and 1-c show the graph created after pruning in sparse areas and in crowded areas

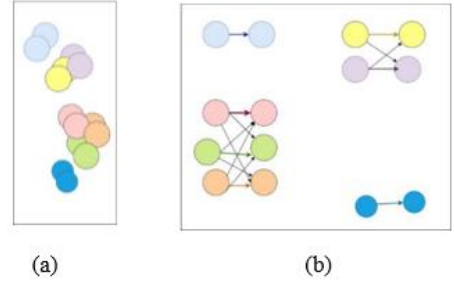

Fig.2. Figure 2-a shows two consecutive frames in one box and Figure 2-b shows the pruned graph made between these two frames

Faster R-CNN architecture is made up of two parts. The first part is a region proposal network. It prepares the Regions of Interest (RoI) with a high chance of existence of an object in them. The second part extracts the RoI features using the RoIPool. Then these features are fed into a classification and regression part [18]. Mask R-CNN network has a similar network to the Faster R-CNN but instead of using RoIPool it uses RoIAlign which leads to higher accuracy. The reason is that RoIAlign uses bilinear interpolation instead of rounding up to map a region to its corresponding feature map.

### B. Defining the Graph

The tracking problem is modeled as a mathematical graph solving problem. The graph is given the input frames and the objects extracted from them. Between every two frames, a bipartite graph is generated in a way that all the objects in the current frame are connected to all the objects in its following frame. Since the graph is bipartite, there shouldn't be any connection between the objects in one frame. Also between two objects in two consecutive frames, there should be only one connecting edge. So the input to the proposed algorithm as a graph is $G = (V, E, \omega)$, where V, E, and $\omega$ denote the set of nodes, the set of edges and weights of edges. The set V is divided into $f$ disjoint sets, each representing one frame and the nodes in each set represents the objects, which are vehicles in the current problem. If $C_i$ where $i \in \zeta: 1 \le i \le f$ shows the $i$-th frame and $v_m^i$ denotes the $m$-th node in the $i$-th frame, thus $C_i = \{V_1^i, V_2^i, V_3^i, ...\}$ and the set of edges can be defined as $E = \{(v_m^i, v_n^j) | n \ne m\}$. The problem with this graph is that it is so crowded which causes massive computations. So it's necessary to prune the graph. Due to small changes in objects position in two consecutive frames, one object can be connected only to its near objects. Thus, the Intersection Over Union (IOU) metric has been used. IOU is a metric to evaluate how well a bounding box and its corresponding ground-truth are fit and is used in evaluating object detection algorithms. By considering the area of bounding box $a$ as $Area(a)$, the metric is defined using (1).

$$IOU(a, b) = \frac{Area(a) \cap Area(b)}{Area(a) \cup Area(b)} \quad (1)$$

So the nodes in a frame are only connected to the nodes in the next frame where the IOU between their bounding boxes is higher than a special threshold. So the set of edges can be changed to (2).

$$E = \{(v_m^i, v_n^i) | n \ne m \text{ and } IOU(i, j) > 0.6\} \quad (2)$$

After pruning, the graph is divided into disjoint graphs, which are either one-by-one graphs or very small graphs. This makes computing costs as little as possible. Fig 1 shows the graph before and after pruning. Fig 2-a shows the overlapping of two consecutive frames in one box to make it more clear how the IOU filtering will perform and Fig 2-b shows the graph achieved from these frames. As it is obvious some of the created graphs contain only two nodes, connected through a single edge, which means some of the tracklets have been created through pruning.

In order to define the appearance model, two different approaches have been used. The first approach utilizes the traditional features of each object which are SIFT and color histogram. However, the appearance model in the second approach uses the features extracted from Mask R-CNN. The following sections will demonstrate how the appearance model through these features is defined.

*1) Appearance Model using SIFT and Color Histogram:*

The Scale Invariant Feature Transform (SIFT) algorithm was pronounced by David G. Lowe in [19] and is considered as a robust algorithm in order to extract image features. To extract these features, multiple steps are performed as follows:

 *1.1 Scale-space extrema detection:* Probable location of features
 *1.2 Key-point localization:* The Accurate location of feature key-points
 *1.3 Orientation assignment:* Assigning orientation to key-points
 *1.4 Key-point descriptor:* Describing key-points as high dimensional vectors

These features are robust against the scale and rotation changes of an image and also perform very well under illumination changes. The algorithm can also handle viewpoint changes to some extent. Also, this algorithm works well on small and highly occluded objects. So it is one of the best algorithms in object re-identification fields. The only problem of this algorithm is that it uses a monochrome intensity image which makes it have the same output of two same images from different colors. So to make the features more discriminative, a color-based feature can be used alongside.

The color histogram of an image provides information about the color distribution of that image. This feature is actually the number of pixels in a specific color. Because color histogram is about the distribution of the colors, two different images of the same distribution may have the same color histogram. Thus this feature per se is not robust for object re-identification but can be combined by SIFT features to discriminate the color features too.

In order to compare the SIFT features, a KNN[6] algorithm is used. The output is the number of matching key-points between the objects $v_m^i$ and $v_n^j$ which is represented by $n_{ij}$. To compare their color histograms, the union between them is calculated as in (3), and finally, the appearance model between objects $v_m^i$ and $v_n^j$ is defined in (4).

$$I(v_m^i, v_n^j) = \frac{\Sigma_{a=1}^n \min(hist_{m,a}^i, hist_{n,a}^j)}{\max(\Sigma_{a=1}^n hist_{m,a}^i, \Sigma_{a=1}^n hist_{m,a}^j)} \quad (3)$$

$$appearance(v_m^i, v_n^j) = n_{ij} \times I(v_m^i, v_n^j) \quad (4)$$

*2) Appearance Model using Deep Network Features:*

To reduce the computing cost of feature extracting, in this method, the feature, extracted from the Mask R-CNN network is used to make the appearance model. According to the architecture of the Mask R-CNN, the features are extracted before the RoIAlign layer. The feature extracted in this network is a 7*7*256 vector of integers. As this vector is too long and consumes very high computational capacity, by using PCA the feature vector size is reduced to 0.1 of the main size. To decide how two nodes are similar, the similarity of the feature vector of these nodes should be calculated. In the proposed method the similarity of two nodes is calculated through cosine similarity metric. Let $F_m^i$ and $F_n^j$ be the feature vectors of $v_n^j$ and $v_m^i$ so the similarity function is defined as (5)

$$similarity(v_m^a, v_n^b) = \frac{F_m^i \cdot F_n^j}{\|F_m^i\| \times \|F_n^j\|} = \frac{\Sigma_{a=1}^n f_{m,n}^i \times f_{n,a}^j}{\sqrt{\Sigma_{a=1}^n (f_{m,a}^i)^2} \times \sqrt{\Sigma_{a=1}^n (f_{n,a}^i)^2}} \quad (5)$$

and the appearance model is defined as (6).

$$appearance(v_m^i, v_n^j) = similarity(v_m^i, v_n^j) \quad (6)$$

*C. Motion Model*

Using a motion model will increase the accuracy of the position estimated in future frames. To have less computational overload, the IOU metric can be used as the motion model too. So the higher the IOU between two nodes is, these nodes are more probable to be the same. Thus, the motion model between the nodes $v_m^i$ and $v_n^j$ is defined using (7).

$$motion(v_m^i, v_n^j) = IOU(v_m^i, v_n^j) \quad (7)$$

the weight of the edge between $v_m^i$ and $v_n^j$ is defined using the appearance model and motion model as in (8). The parameters $\alpha$ and $\beta$ are the weights of each element that are assigned by examining different values.

$$\omega(v_m^i, v_n^j) = \alpha(motion(v_m^i, v_n^j)) + \beta(appearance(v_m^i, v_n^j)) \quad (8)$$

*D. Handling Occlusion and a leaving node*

One of the challenges that is very probable in multi-object tracking is occlusion. The occlusion refers to the non-existence of an object in a scene for a while. It usually happens because of hiding behind another object or an obstacle, leaving the scene or even malfunctioning of the object's detector. In the proposed method to handle the occlusion, a state model has been used. Besides to manage the missing objects a hypothetical node has been replaced. Each node can be in three states: tracking, lost and left. Every node which enters the graph will be in tracking state and remains in this tracking unless being lost. So if one node is missing in a frame it means that for some reason it has left the scene or it has been occluded by another object or the detector missed it. To determine if the node has been lost or not, the last place where the node has been monitored is considered. If the node has been lost near the borders of the frame it's very probable that the node has left the scene and the state of the node changes to left otherwise, it has been occluded and the nodes are in a lost state. If a node is in the lost state then it will be replaced with a hypothetical node. This node, like the real node, will have an appearance and motion model and as it's used instead of the real node, it should have feature vector, position, motion model and appearance model. The feature vector and appearance model of the hypothetical node are the same as the original node but the position should be estimated. To estimate the new position of the hypothetical node, the average speed of the node is calculated using (9) where $pos(v)$ is the position of a node.

---
[6] K-Nearest Neighbor

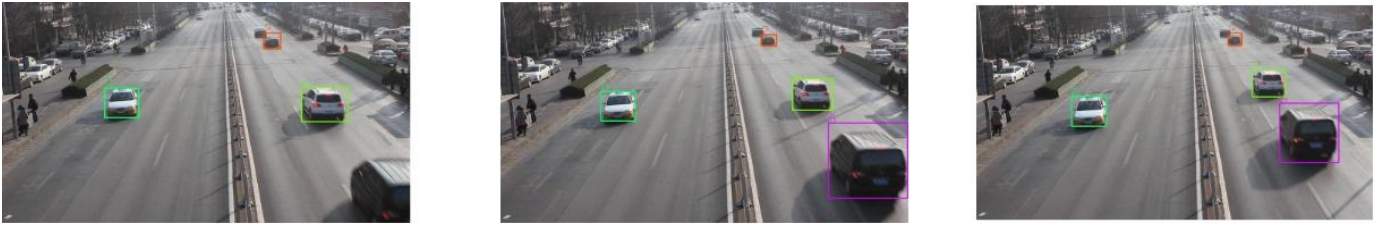

Fig.3. The tracking result on UA-DETRAC benchmark. The cars in bounding boxes with same color belong two same trajectory

$$velocity(v_m^i) = \frac{pos(v_{m-1}^i) - pos(v_5^i)}{\frac{m-5}{fps}} \quad . \quad (9)$$

The *fps* is the video rate in frame per second and should be considered in calculating the velocity of the moving object. After calculating the velocity the approximate position is calculated using (10)

$$pos(v_m^i) = pos(v_{m-1}^i) + velocity(v_{m-1}) \times fps \quad (10)$$

## IV. EVALUATION AND IMPLEMENTATION

To implement the proposed method, The UA-DETRAC dataset has been used. This dataset is introduced in [20] and consists of 140,000 frames captured at the rate of 25 frames per second with a resolution of $960 * 540$ having the total number of 8,250 vehicles over all frames. To extract objects and their corresponding features, the Mask R-CNN network is being used. The total project is implemented by Python and the program was run on google Colab to extend the project's speed. Fig 3 shows the result of the Algorithm on the frames of UA-DETRAC.

Since multiple-object tracking is used widely in different fields, there are a lot of datasets designed for this purpose. Because the issue here has been reduced to track cars we can use the UA-DETRACK dataset to test the method.

The overall results of these systems can be seen in TABLE I. The results announced are the average results from executing the method on different detection confidence. So the impact of the detection confidence on the tracking result has vanished. The metrics used in the evaluation are clear-MOT metrics, which are described in [21]. These metrics are multi-object tracking accuracy (MOTA), false negative (FN), false positive (FP) and ID switch (IDSW).

The FP is the number of tracker's outputs which do not exist in the ground-truth, the FN is the number of the objects in a trajectory missed by the tracker, the IDSW is the number of times that an objects ID changes and MOTA is the accuracy of a tracker calculated through (11).

$$MOTA = 1 - \frac{\Sigma_t(FN_t + FP_t + IDSW_t)}{\Sigma_t GT_t} \quad (11)$$

The results obtained from evaluating the two proposed methods are mentioned in TABLE I. According to these features, using SIFT alongside with color histogram outperforms using the features from the Mask R-CNN. The reason is that SIFT features are known as robust features widely used in re-identification fields but the features extracted from deep networks are mainly used in the categorizing of each object. Thus they couldn't have good performance in this tracking field. Both IDSW and FP numbers in the first method are significantly lower than the second method. These two metrics depend on how good the appearance model performs which in this case depends on the feature vectors. So it can be concluded that the SIFT features are more robust than the deep features in this case. The number of FNs mainly depends on the detector result. And it is why the two proposed methods resulted in approximately the same numbers. The number of FN in the first method is higher than the second method which is the cause of the SIFT comparing threshold. Furthermore, these proposed methods are compared with some famous tracking methods published in the UA-DETRAC dataset. In comparison with the IOU tracker, the SIFT method had a better performance. The reason is that the IOU tracker only relies on the IOU threshold which may not have a good performance in crowded areas.

## V. CONCLUSION

In this paper, an efficient novel method for multiple-object tracking is proposed. For this purpose, two different methods have been suggested to extract object features. The first approach uses SIFT and color histogram features of each image while the second one uses the deep features obtained from the object detector network. To find the matching objects in different frames, the problem is modeled as a graph best route problem. In addition, for minimizing the computational costs, the graph has been pruned using IOU filters.

According to the results, the first approach significantly outperforms the second method's accuracy. The reason is that the features used in the first approach are more significant in re-identification tasks while the deep features used in the second approach are more accurate in categorization tasks that couldn't perform well in the tracking problem but has less computational cost than the first approach.

To improve the method proposed here, one can use the low computational cost of more significant neural network features which will offer more accurate features on re-identification tasks.

TABLE I. EVALUATING THE PROPOSED METHODS ON UA-DETRAC DATASET. ALL RESULTS ARE AVERAGED OVER DIFFERENT INPUT DETECTION CONFIDENCE THRESHOLD

| Methods | Metrics | | | |
|---|---|---|---|---|
| | MOTA | FP | FN | IDSW |
| Proposed Method + SIFT | 58.9 | 244 | 1898 | 55 |
| Proposed Method + Deep Features | 15.9 | 13021 | 1637 | 1637 |
| IOU [7] | 19.4 | 14796 | 171806 | 2311 |
| CEM [22] | 5.1 | 12341 | 260390 | 267 |
| H$^2$T [5] | 12.4 | 148 | 51765 | 173899 |
| CMOT [23] | 12.6 | 32093 | 180184 | 285 |
| GOG [24] | 14.2 | 32093 | 180184 | 3335 |

REFERENCES

[1] A. Yilmaz, O. Javed, and M. Shah, "*Object tracking: A survey,*" Acm computing surveys (CSUR), 2006. 38(4): p. 13-es.
[2] X. Li, et al, "*A survey of appearance models in visual object tracking,*" ACM transactions on Intelligent Systems and Technology (TIST), 2013. 4(4): p. 1-48.
[3] J. Berclaz, et al., "*Multiple object tracking using k-shortest paths optimization,*" IEEE transactions on pattern analysis and machine intelligence, 2011. 33(9): p. 1806-1819.
[4] Z. Xi, et al., "*Multiple object tracking using the shortest path faster association algorithm,*" The Scientific World Journal, 2014. 2014.
[5] L. Wen, et al. "*Multiple target tracking based on undirected hierarchical relation hypergraph*, " in *Proceedings of the IEEE conference on computer vision and pattern recognition*. 2014.
[6] A. R. Zamir, A. Dehghan, and M. Shah. "*Gmcp-tracker: Global multi-object tracking using generalized minimum clique graphs*, " in *European Conference on Computer Vision*. 2012. Springer.
[7] E. Bochinski, V. Eiselein, and T. Sikora. "*High-speed tracking-by-detection without using image information*," in *2017 14th IEEE International Conference on Advanced Video and Signal Based Surveillance (AVSS)*. 2017. IEEE.
[8] E. Bochinski, T. Senst, and T. Sikora. "*Extending IOU based multi-object tracking by visual information*, " in the *2018 15th IEEE International Conference on Advanced Video and Signal Based Surveillance (AVSS)*. 2018. IEEE.
[9] W. Choi, and S. Savarese. "*A unified framework for multi-target tracking and collective activity recognition,*" in *European Conference on Computer Vision*," 2012. Springer.
[10] K. Mu, F. Hui, and X. Zhao, "*Multiple Vehicle Detection and Tracking in Highway Traffic Surveillance Video Based on SIFT Feature Matching,*" Journal of Information Processing Systems, 2016. 12(2).
[11] G. Ning, et al, "*Spatially supervised recurrent convolutional neural networks for visual object tracking,*" in the *2017 IEEE International Symposium on Circuits and Systems (ISCAS)*. 2017. IEEE.
[12] M. Zhai, et al, "*Deep learning of appearance models for online object tracking,*" in *Proceedings of the European Conference on Computer Vision (ECCV)*. 2018.
[13] L. Bertinetto, et al, "*Fully-convolutional siamese networks for object tracking,*" in the *European conference on computer vision*. 2016. Springer.
[14] A. Bewley, et al, "*Simple online and realtime tracking,*" in *2016 IEEE International Conference on Image Processing (ICIP)*. 2016. IEEE.
[15] N. Wojke, A. Bewley, and D. Paulus, "*Simple online and realtime tracking with a deep association metric,*" in the *2017 IEEE international conference on image processing (ICIP)*. 2017. IEEE.
[16] X. Zhang, et al, "*Multi-target tracking of surveillance video with differential YOLO and DeepSort,*" in *Eleventh International Conference on Digital Image Processing (ICDIP 2019)*. 2019. International Society for Optics and Photonics.
[17] X. Hou, Y. Wang, and L. -P. Chau,. "*Vehicle Tracking Using Deep SORT with Low Confidence Track Filtering*," in *2019 16th IEEE International Conference on Advanced Video and Signal Based Surveillance (AVSS)*. 2019. IEEE.
[18] S. Ren, et al. *Faster r-cnn: Towards real-time object detection with region proposal networks,*" in *Advances in neural information processing systems*. 2015.
[19] D. G. Lowe, "*Distinctive image features from scale-invariant keypoints,*" International journal of computer vision, 2004. 60(2): p. 91-110.
[20] L. Wen, et al., "*UA-DETRAC: A new benchmark and protocol for multi-object detection and tracking,*" Computer Vision and Image Understanding, 2020: p. 102907.
[21] K. Bernardin and R. Stiefelhagen, "*Evaluating multiple object tracking performance: the CLEAR MOT metrics,*" EURASIP Journal on Image and Video Processing, 2008. 2008: p. 1-10.
[22] A. Milan, S. Roth, and K. Schindler, "*Continuous energy minimization for multitarget tracking,*" IEEE transactions on pattern analysis and machine intelligence, 2013. 36(1): p. 58-72.
[23] S. -H. Bae and K. -J. Yoon, "*Confidence-based data association and discriminative deep appearance learning for robust online multi-object tracking,*" IEEE transactions on pattern analysis and machine intelligence, 2017. 40(3): p. 595-610.
[24] H. Pirsiavash, D. Ramanan, and C. C. Fowlkes "*Globally-optimal greedy algorithms for tracking a variable number of objects*. in *CVPR 2011*. 2011. IEEE.